\RequirePackage{fix-cm}
\documentclass[12pt,a4paper]{article}
\makeatletter\if@twocolumn\PassOptionsToPackage{switch}{lineno}\else\fi\makeatother

\usepackage{amsfonts,amssymb,amsbsy,tabulary,amsmath}
\usepackage{amsmath,tabulary,graphicx,fancyhdr,times}
\usepackage[utf8]{inputenc}
\usepackage{helvet}

\usepackage{abstract}
\AtBeginDocument{\usepackage{lastpage}}
\usepackage{caption}
\def\NormalBaseline{\def\baselinestretch{1.1}}
\usepackage{textcase} 
\usepackage[T1]{fontenc}
\usepackage[margin=2cm,top=30mm,bottom=1in,columnsep=20pt]{geometry}
 \date{}
\renewenvironment{abstract}
{\vspace*{-1pc}\trivlist\item[]\leftskip\oupIndent\hrulefill\par\vskip4pt\noindent{\bfseries Abstract\par\noindent\\\noindent}\hspace*{10pt}\fontsize{9}{10.8}\selectfont}{\par\noindent\hrulefill\endtrivlist}
\makeatletter\def\oupIndent{1pt}
\def\author#1{\gdef\@author{\hskip-\dimexpr(\tabcolsep)\hskip\oupIndent\parbox{\dimexpr\textwidth-\oupIndent}{\centering\fontsize{12}{14.4}\selectfont#1}}}
\def\title#1{\gdef\@title{\centering\def\baselinestretch{1}\selectfont\ifx\@articleType\@empty\else\@articleType\\\fi\fontsize{16}{19.2}\selectfont#1}}
\fancypagestyle{headings}{\fancyhf{}\fancyhead[C]{\RunningHead}\fancyhead[R]{\thepage}}

\fancypagestyle{plain}{\fancyhf{}\fancyhead[C]{\RunningHead}\fancyhead[R]{\thepage}\fancyfoot[L]{}}
\let\@articleType\@empty \def\articletype#1{\gdef\@articleType{{\itshape#1}}}
\usepackage[nobottomtitles*]{titlesec}
\titleformat{\section}[hang]{\NormalBaseline\raggedright\large\bfseries\boldmath\fontsize{14}{16.8}\selectfont}
{\thesection.\hspace{-6pt}}
{10pt}
{}
[]
\titleformat{\subsection}[hang]{\NormalBaseline\filright\itshape\fontsize{14}{16.8}\selectfont}
{\thesubsection\hspace{-6pt}}
{10pt}
{}
[]
\titleformat{\subsubsection}[hang]{\NormalBaseline\filright\itshape\fontsize{12}{14.4}\selectfont}
{\thesubsubsection\hspace{-6pt}}
{10pt}
{}
[]
\titleformat{\paragraph}[runin]{\NormalBaseline\filright\itshape\fontsize{12}{14.4}\selectfont}
{\theparagraph\hspace{-6pt}}
{10pt}
{}
[]
\titleformat{\subparagraph}[runin]{\NormalBaseline\filright\fontsize{12}{14.4}\selectfont}
{\thesubparagraph\hspace{-6pt}}
{10pt}
{}
[]

\titlespacing{\section}{0pt}{1.5\baselineskip}{.2\baselineskip}  
\titlespacing{\subsection}{0pt}{1\baselineskip}{.2\baselineskip}  
\titlespacing{\subsubsection}{0pt}{1.5\baselineskip}{.2\baselineskip}  
\titlespacing{\paragraph}{0pt}{.5\baselineskip}{10pt}  
\titlespacing{\subparagraph}{0pt}{.5\baselineskip}{10pt} 
\makeatother

\usepackage{url,multirow,morefloats,floatflt,cancel} 
\makeatletter

\AtBeginDocument{\@ifpackageloaded{textcomp}{}{\usepackage{textcomp}}}
\makeatother
\usepackage{colortbl}
\usepackage{xcolor}
\usepackage{pifont}
\usepackage[nointegrals]{wasysym}

\usepackage[disable]{todonotes}%

\usepackage{mathptmx}
\usepackage{graphics}
\usepackage{graphicx}
\usepackage{adjustbox}
\usepackage{varwidth}

\usepackage{amsmath}
\usepackage{comment}

\usepackage{caption}
\usepackage{subcaption}
\usepackage{times}
\usepackage{color}				
\usepackage{xifthen}				
\usepackage{amsmath}
\usepackage{amssymb}
\usepackage{psfrag}
\usepackage{epstopdf}
\usepackage{ulem}
\usepackage{balance}
\usepackage{booktabs}

\usepackage[affil-it]{authblk}

\usepackage{algpseudocode,algorithm,algorithmicx}

\newcommand*\Let[2]{\State #1 $\gets$ #2}
\algrenewcommand\algorithmicrequire{\textbf{Precondition:}}
\algrenewcommand\algorithmicensure{\textbf{Postcondition:}}

\usepackage{bm}

\DeclareMathOperator*{\argmin}{arg\,min}
\DeclareMathOperator*{\argmax}{arg\,max}

\makeatletter

\def\mcWidth#1{\csname TY@F#1\endcsname+\tabcolsep}

\def\cAlignHack{\rightskip\@flushglue\leftskip\@flushglue\parindent\z@\parfillskip\z@skip}
\def\rAlignHack{\rightskip\z@skip\leftskip\@flushglue \parindent\z@\parfillskip\z@skip}

\usepackage{ifxetex}
\ifxetex\else\if@twocolumn\usepackage{dblfloatfix}\fi\fi

\AtBeginDocument{
\expandafter\ifx\csname eqalign\endcsname\relax
\def\eqalign#1{\null\vcenter{\def\\{\cr}\openup\jot\m@th
  \ialign{\strut$\displaystyle{##}$\hfil&$\displaystyle{{}##}$\hfil
      \crcr#1\crcr}}\,}
\fi
}

\AtBeginDocument{%
  \@ifpackageloaded{endfloat}%
   {\renewcommand\efloat@iwrite[1]{\immediate\expandafter\protected@write\csname efloat@post#1\endcsname{}}}{\newif\ifefloat@tables}%
}%

\def\BreakURLText#1{\@tfor\brk@tempa:=#1\do{\brk@tempa\hskip0pt}}
\let\lt=<
\let\gt=>
\def\processVert{\ifmmode|\else\textbar\fi}

\@ifundefined{subparagraph}{
\def\subparagraph{\@startsection{paragraph}{5}{2\parindent}{0ex plus 0.1ex minus 0.1ex}%
{0ex}{\normalfont\small\itshape}}%
}{}

\newcommand\role[1]{\unskip}
\newcommand\aucollab[1]{\unskip}
  
\@ifundefined{tsGraphicsScaleX}{\gdef\tsGraphicsScaleX{1}}{}
\@ifundefined{tsGraphicsScaleY}{\gdef\tsGraphicsScaleY{.9}}{}
\def\checkGraphicsWidth{\ifdim\Gin@nat@width>\linewidth
	\tsGraphicsScaleX\linewidth\else\Gin@nat@width\fi}

\def\checkGraphicsHeight{\ifdim\Gin@nat@height>.9\textheight
	\tsGraphicsScaleY\textheight\else\Gin@nat@height\fi}

\def\fixFloatSize#1{}
\let\ts@includegraphics\includegraphics

\def\inlinegraphic[#1]#2{{\edef\@tempa{#1}\edef\baseline@shift{\ifx\@tempa\@empty0\else#1\fi}\edef\tempZ{\the\numexpr(\numexpr(\baseline@shift*\f@size/100))}\protect\raisebox{\tempZ pt}{\ts@includegraphics{#2}}}}

\AtBeginDocument{\def\includegraphics{\@ifnextchar[{\ts@includegraphics}{\ts@includegraphics[width=\checkGraphicsWidth,height=\checkGraphicsHeight,keepaspectratio]}}}

\DeclareMathAlphabet{\mathpzc}{OT1}{pzc}{m}{it}

\def\URL#1#2{\@ifundefined{href}{#2}{\href{#1}{#2}}}

\def\UrlOrds{\do\*\do\-\do\~\do\'\do\"\do\-}%
\g@addto@macro{\UrlBreaks}{\UrlOrds}

\edef\fntEncoding{\f@encoding}

\makeatother

\newif\ifmultipleabstract\multipleabstractfalse%
\usepackage[numbers,sort&compress,super]{natbib}

\def\ie{i.\,e.}

\def\bB{\mathbf{B}}
\def\bD{\mathbf{D}}
\def\bK{\mathbf{K}}
\def\bI{\mathbf{I}}
\def\bJ{\mathbf{J}}
\def\bR{\mathbf{R}}
\def\bu{\mathbf{u}}

\def\bg{\mathbf{g}}
\def\bC{\mathbf{C}}

\def\vtopa{v\rightarrow p}

\newcommand{\showComments}		{1}

\newcommand{\comments}[2]
{\ifthenelse{\equal{1}{#1}}{{#2}}{}}
\newcommand{\changes}[3]
{
  \ifthenelse{\equal{1}{#1}}
    {{ #2}}
    {{ #3}}
}

\newcommand{\colorJG}[1]{{\color[rgb]{0.0,0.0,0.0} {#1}}}
\newcommand{\colorRW}[1]{{\color[rgb]{0.0,0.0,0.0} {#1}}} 
\newcommand{\colorFIX}[1]{{\color[rgb]{0.0,0.0,0.0} {#1}}} 

\newcommand{\colorCorrection}[1]{{\color[rgb]{0,0,0.0} {#1}}}
\newcommand{\colorCorrectionB}[1]{{\color[rgb]{0,0,0.0} {#1}}}

\newcommand{\JG}[1]{\comments{\showComments}{{\colorJG{#1}}}}

\newcommand{\RW}[1]{\comments{\showComments}{{\colorRW{#1}}}} 
\newcommand{\FIX}[1]{\comments{\showComments}{{\colorFIX{#1}}}} 

\newcommand{\CO}[1]{\comments{\showComments}{{\colorCorrection{#1}}}}
\newcommand{\CoB}[1]{\comments{\showComments}{{\colorCorrectionB{#1}}}}

\def\metric{$\mu\pm\sigma$(max)}

\newcommand\improvedGPR{iGPR}
\newcommand\initializedFEM{BGM}

\newcommand\vesselGM{VCGM}
\newcommand\adaptGM{ACGM}
\newcommand\fineFEM{fineBGM}

\newcommand{\grph}{\mathcal{G}}

\newcommand{\comp}{\bC_{\pi_{t}}}  
\newcommand{\mFEM}{\mathcal{T}_{\pi_{t}}}

\newcommand{\PotCand}{\mathcal{P}}
\newcommand{\matchFew}{\mathbf{P}}

\newcommand{\MCpt}{MC_{\pi_{t}}}

\newcommand{\thMC}{$MC_{TH}$}

\newcommand{\thGEO}{G_{TH}}
\newcommand{\tMGEO}{\mathbf{G}_{t}}
\newcommand{\sMGEO}{\mathbf{G}_{s}}

\newcommand{\sBfree}{\vec{X}^{SF}}
\newcommand{\tBfree}{\vec{X}^{TF}}
\newcommand{\sBif}{\mathbf{s}}
\newcommand{\tBif}{\mathbf{t}}
\newcommand{\sBifMax}{\mathbf{s}_{maxD}}
\newcommand{\tBifMax}{\mathbf{t}_{maxD}}
\newcommand{\sBifN}{$|\vec{X}^S|$}
\newcommand{\tBifN}{$|\vec{X}^T|$}

\usepackage[T1]{fontenc}
\begin{document}

\title{Elastic registration based on compliance analysis and biomechanical graph matching}
\author{Jaime Garcia Guevara$^{1,2}$  
	\thanks{Electronic address: \texttt{jaime.garcia-guevara@inria.fr}; Corresponding author}
	}
\author{Igor Peterlik$^1$ }
\author{Marie-Odile Berger$^{1,2}$ }
\author{Stephane Cotin$^1$ \fontsize{11.5}{13.8}{}\fontsize{8}{9.6}\itshape\selectfont{}}

\affil{$^1$Inria Nancy Grand Est, Villers-les-Nancy, France \\ 
	$^2$Universit\'e de Lorraine, Nancy, France}	

\def\RunningHead{{~}}
\maketitle 

\begin{abstract}
An automatic elastic registration method suited for vascularized organs is proposed. The vasculature in both the preoperative and intra-operative images is represented as a graph. A typical application of this method is the fusion of pre-operative information onto the organ during surgery, to compensate for the limited details provided by the intra-operative imaging modality (e.g. CBCT) and to cope with changes in the shape of the organ. Due to image modalities differences and organ deformation, each graph has a different topology and shape.
\CoB{The Adaptive Compliance Graph Matching (\adaptGM{}) method} presented does not require any manual initialization, handles intra-operative nonrigid deformations of up to 65\,mm and computes a complete displacement field over the organ from only the matched vasculature.
\CoB{\adaptGM{} is better than the previous Biomechanical Graph Matching method\cite{GarciaGuevara2018} (\initializedFEM{}) because it} uses an efficient biomechanical vascularized liver model to compute the organ's transformation and the vessels bifurcations compliance.
\CoB{This allows to efficiently find the best graph matches with a novel compliance-based adaptive search.} 
These contributions are evaluated on ten realistic synthetic and two real porcine automatically segmented datasets. 
\adaptGM{} obtains better target registration error (TRE) \CoB{than \initializedFEM{}, with an average TRE in the real datasets of 4.2 mm compared to 6.5 mm, respectively.} It also is up to one order of magnitude faster, less dependent on the parameters used and more robust to noise. 
\end{abstract}\textbf{Index terms: }{Non-rigid registration, biomechanics, data fusion, graph matching} 
    
\section{Introduction}
\label{s:intro}

Providing enhanced visualization (e.g. of an organ internal structures) during an intervention can significantly improve surgical procedures. Since each image modality provides different and often complementary information on tissue structures or deformation changes, the fusion of intra-operative and preoperative images into a unique coordinate frame adds significant value \cite{Peters2008}. Registration of the preoperative image onto an intra-operative image (X-ray, Ultra Sound, Cone Beam CT, etc) can be handled in many different manners, given the application, the image modality, the parameter space, or the optimization process. The literature is vast on the topic and several surveys classify the different methods for this problem\cite{sotiras2013deformable}. 
Regardless of the targeted clinical application, this paper focuses on vessel based registration approaches, with special interest into the type of deformation that is handled, the way the global displacement field is reconstructed, and their compatibility with the clinical work flow.

Feature-based registration methods require an identification of correspondences between the pre-operative and intra-operative images. Being present in many anatomical structures, vessels can be used as features for multi-modal image registration. As intrinsic and natural features, they eliminate the need for markers, making this solution more compatible with clinical constraints. More generally, graph-like structures are common in medical images and can be obtained from angiography or airway trees, to mention only some of them. Robustly registering such graphs is thus a key enabling technology for preoperative planing, intra-operative navigation, follow-up or group-wise analysis \cite{Matl2017}. However, the topology and shape of the vascular graphs is not consistent between patients \cite{Groher2007}, and, when considering intra-patient registration, variations in image quality also lead to very different graphs to be matched, making this problem quite challenging.

Several  methods use euclidean or geodesic metrics of nodes to match graphs.
Often, root nodes are manually matched \cite{xue2006}, or a global rigid or affine alignment is first performed before using fine graph-matching techniques \cite{MoriconiElasticGeo2018}. 
These methods encode nodes and edges similarity in an affinity matrix $A$. Graph matching is then formulated as a quadratic assignment problem which maximizes a global consistency score based on $A$ \cite{LeordeanuSpectral2005,ChoRRW_GM2010,Zhou2015FactorizedGM}.
However, defining a node- or edge-similarity metric is not easy due to the non-linear deformation which may occur between the two acquisitions. In addition, due to segmentation inaccuracies, geodesic constraints may not be exactly satisfied. Topology changes may also appear.
When graph nodes and edges don't have distinctive features and only the vessel geometry is available, the methods based on local similarities are ineffective. 
The use of over-connected graph has thus been proposed\cite{MoriconiElasticGeo2018}. 
This approach incorporates in the graph the edges between nodes connected by vessels within a neighborhood with a given radius. This compensates for topological inaccuracies or deformation of branches. However, robustness to large deformation is not guaranteed since nothing can ensure that the established correspondences are physically coherent with the elasticity of the organ. In addition, the process is dependent from the initial rigid registration.

Within a intra-operative context, the transformation model needs to efficiently and accurately describe high deformation anatomical properties \cite{sotiras2013deformable}.
Computational efficiency has been achieved with simple models as thin-plate splines transformations \cite{Lange20093D}, however this deformation is not necessarily physically coherent.
Seradell \textit{et al.} \cite{Serradell2015NonRigid} proposed an approach which does not rely on local similarity but uses a deformation model to determine whether a new pair of matching is compatible with the current set of matching hypotheses. 
Using Gaussian Process Regression (GPR), this method iteratively generates hypotheses while the search space is refined.
The search space defines the most likely correspondences to explore at every step.
On several examples and without initialization, this method handled topological differences and deformation.
Nonetheless, the GPR mapping can not handle large non-linear deformations and thus it may only find incomplete correspondences.
Moreover, the matching time with large graphs does not satisfy intraoperative constraints.
To efficiently match large graphs, a Monte Carlo tree search method was proposed \cite{pinheiro2017}. 
It explores new matches and extends good matches in a balanced way.
Its efficiency relies in a path descriptor and an implicit transformation model that assume a rigid transformation with small nonlinear deformation. Thus, this method is not suitable for large deformations.

Finally, in order to have a more discriminative deformation model without a prohibitive computational cost at each iteration, Pinheiro \textit{et al} \cite{Pinheiro-ISBI2018} recently proposed to use a B-spline deformation model which is updated incrementally using a Kalman filter. 
While this process saves time, B-splines are not able to properly describe elastic properties of the organs.
Generally speaking, the choice of a constitutive model is essential for the registration process when considering deformable organs. 
From the previous methods \cite{Lange20093D,Serradell2015NonRigid,pinheiro2017,Pinheiro-ISBI2018}, the models used reach computational efficiency. 
However, physical consistency with organ properties, in particular heterogeneities and anisotropy of the tissues, is not guaranteed.
To this end, more realistic, biomechanically-driven models have been proposed to solve elastic image intensity based registration problems \cite{oktay2013biomech}. However, the combination of graph matching and biomechanics is a new direction.

A biomechanical graph matching method (\initializedFEM{}) which combines a GPR approach with a biomechanical model of the organ, as a mean to discard matching hypotheses which would lead to non-plausible deformations was proposed ~\cite{GarciaGuevara2018}. 
However, just replacing the GPR by a biomechanical model is extremely costly in terms of computational time, as the physics-based simulation takes about three orders of magnitude longer than GPR. 
\initializedFEM{} is initialized with a GPR solution to reduce the matching time. 
This allowed to recover additional matches compatible with the elasticity of the organ even when large elastic deformations were considered.
However, it was not robust to noise, only matched limited size graphs and the computation time was still above intra-operative constraints. 


Several contributions with respect to \initializedFEM{} are presented in this article: 
i) the use of a more advanced biomechanical model to handle heterogeneities and anisotropy due to the vascularization;
ii) the definition of a better \CoB{and novel} metric for generating improved graph-matching hypotheses, based on the notion of compliance, the inverse of the stiffness. 
iii) the generation of matching hypotheses using an adaptive \CoB{bounded region}.
\CoB{The first two contributions compose the Vessels Compliance Graph Matching method (\vesselGM{}), which reduces the matching search space and/or improves the registration accuracy. Moreover, the three contributions constitute the Adaptive Compliance Graph Matching method (\adaptGM{}), which further reduces the computation time by predicting first the most plausible matching hypotheses and reduces the sensitivity on the search space parameters. These contributions improve the registration quality and meet intra-operative timing constraints.}

\CoB{This article describes the mechanical model used and both novel methods (\vesselGM{} and \adaptGM{}). 
	It also presents experiments on both synthetic and real data, including a sensitivity analysis highlighting the robustness and genericity of the compliance matching methods.}  


\section{Materials and Methods}
\label{s:method}
\CO{Two porcine liver datasets (PA and PB) were acquired, under institutionally approved animal ethics protocol, with contrast agent injection. This article does not contain patient data.} 
\CoB{For each pre- and intra-operative images, the vessels are segmented using a model-based
tube detection filter combined with a region growing algorithm~\cite{Smistad2014GPU}.
Then from each segmentation, Dijkstra minimum-cost spanning trees are recursively constructed to obtain the vessels' graph ~\cite{Plantefeve2017}.}  
Because the fully automatic vessels' segmentation is often inaccurate and incomplete, especially in the case of the intra-operative images, the resulting graphs have important topological differences which must be correctly addressed by the matching algorithm.

\begin{figure}[H]
\captionsetup{belowskip=-15pt}	
\centerline {\adjincludegraphics[width=1.0\textwidth, trim={ 0 0 0 0},clip]{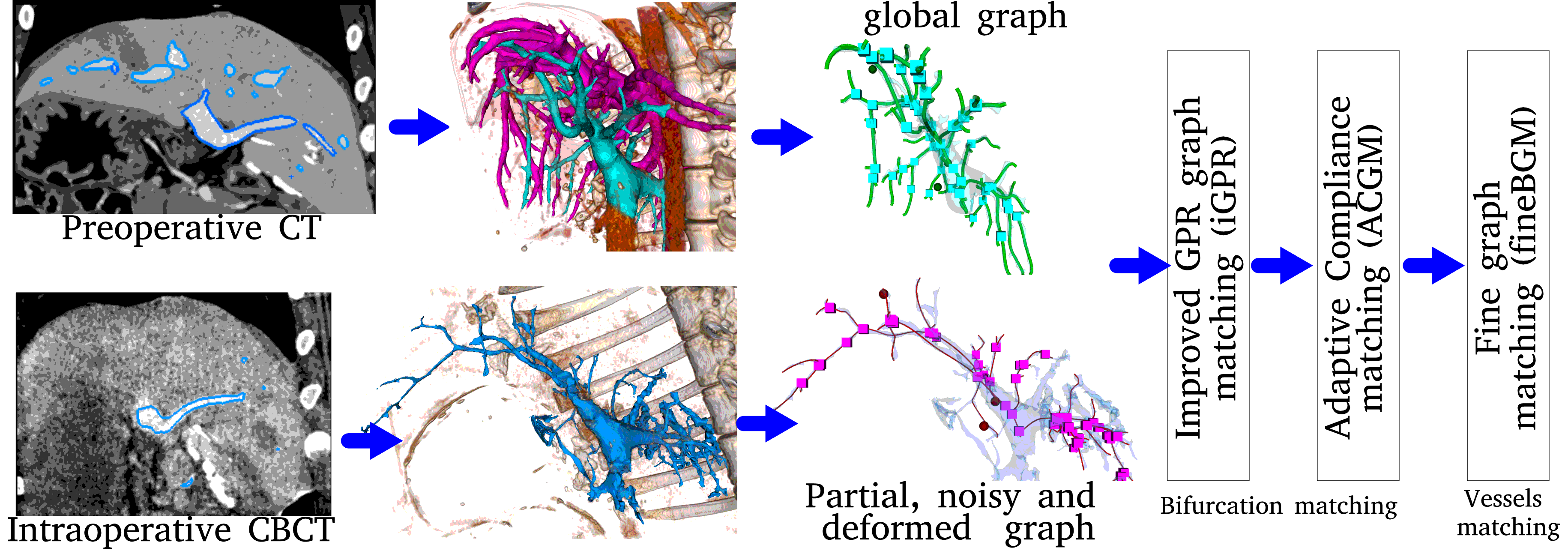}}
\caption{Description of the complete registration pipeline: The graphs are extracted from the vessels segmentation of pre- and intra-operative images (CTA and CBCT in this example). Then, the graphs bifurcations are matched (mostly rigid and incompletely) with \improvedGPR{}. This first matching is used to initialize \adaptGM{}, which finds a complete deformable bifurcations match very efficiently. This compliance-based matching (\adaptGM{}) is the main contribution described in this article. Finally, the fine FEM-based alignment (\fineFEM{}) of the graph edges is performed. } 
\label{fig:pipeline}
\end{figure}

\CoB{The complete registration pipeline}, shown in Fig.~\ref{fig:pipeline}, consists of:
(i) The improved graph matching algorithm (\improvedGPR{}~\cite{GarciaGuevara2018}) that reduces significantly the matching time. 
The result of \improvedGPR{} is usually incomplete, generally misses parts with big deformation. Thus, it is only used as a initialization for the next step.
(ii) \adaptGM{} reduces the search space by using the compliance, while finding most of the bifurcations matches including those with large deformations. 
(iii) The fine alignment of graphs edges using the biomechanical model. 
\CoB{The first and third parts were previously presented in Garcia-Guevara et al.~\cite{GarciaGuevara2018}. While, \adaptGM{} comprises the contributions described in the following sections. }

\subsection{Fast biomechanical model of vascularized liver}
\label{ss:meth:model}

The original GPR matching~\cite{Serradell2015NonRigid} is relatively generic and, depending on the values of hyperparameters, may theoretically adapt to large range of deformations.
Nevertheless, the deformations which occur in soft tissues during surgical manipulations display a high level of non-linearity due to their complex characteristics. Typically, internal structures such as vessels introduce heterogeneity and anisotropy which cannot properly be taken into account without a biomechanical model. In this work, the purpose of the biomechanical model is to compute (i) the transformation of the pre-operative data given the actual hypothesis, and (ii) the compliance which replaces the covariance in the original GPR matching algorithm. 

The biomechanical model considered in this work is based on the co-rotational formulation of linear elasticity~\cite{Nesme2005}. Although this approach relies on a linear stress-strain relationship, it provides a good approximation of large deformations including rotations. Further, a composite FE approach which accounts for the mechanics of both of parenchyma and vessels~\cite{Peterlik2012, plantefeve2016patient} is used to model vascularized tissue. 
The parenchyma is modeled with linear P1 tetrahedral elements where for each element $p$, the local 12$\times$12 stiffness matrix $\bK_p$ is computed as 
\begin{equation}
\bK_p = \bR_p(\bu_p)^\top \left\{ \int_{V_p} \bB_p^\top\bD_p\bB_p dV \right\} \bR_p(\bu_p)
\end{equation}
where $\bB_p$ and $\bD_p$ are, respectively, the strain-displacement and stress-strain matrices which remain constant during the simulation, and $\bR_p$ is a matrix composed of the element rotation matrix which depends on the actual displacement $\bu_p$ of the parenchyma mesh nodes thus introducing a non-linearity into the formulation~\cite{Nesme2005}. \CoB{The vascular structures are modeled as trees composed of serially-linked Timoshenko beam elements, mimicking the biomechanics of hollow tubes parametrized with Young's modulus, diameter and wall thickness.} The beam mechanics includes both positional and rotational degrees of freedom (DoF). Hence, each beam element $v$ is modeled with a 12$\times$12 local stiffness matrix $\bK_v$ which depends on the actual displacements and orientations of the beam nodes~\cite{Przem1985, duriez2006new}.

Despite the identical size, the element matrices $\bK_p$ and $\bK_v$ have a completely different structure: the former describes mechanics of an element given by 4 nodes each determined by 3 positional DoFs, while the latter determine the behavior of element having two nodes, each equipped with 3 positional and 3 rotational DoFs. \CoB{The coupling mechanism~\cite{Peterlik2012} uses a mapping} that defines a Jacobian matrix $\bJ_{\vtopa}$ which is used to compute a composite stiffness matrix $\bK_e$ as 
\begin{equation}
\bK_e = \bK_p + \bJ^\top_{\vtopa} \bK_v \bJ_{\vtopa}.
\end{equation}
\CoB{The respective references detail the generation of the tetrahedral mesh of the parenchyma~\cite{plantefeve2016patient} and beam tree representing the vascular structure~\cite{Plantefeve2017}.}

Given the biomechanical FE model represented by a global stiffness matrix $\bK$ assembled from composite local matrices $\bK_e$, 
the transformation corresponding to a matching hypothesis $\pi$ given by pairs of bifurcations $\sBif_i \leftrightarrow \tBif_j$ is computed as follows.
As certain mesh nodes $n_i$ are generated to coincide with the set of source bifurcations $\vec{X}^S$,
 each bifurcation matching pair determines a non-homogeneous Dirichlet condition that drives the deformation model, thus  $\bu_{n_i} = \tBif_j - \sBif_i$ describes the displacement of a node $n_i$.
A penalty method defines the Dirichlet conditions. This method is physically interpreted as adding very stiff linear springs to each node $n_i$. These springs pull the nodes from its initial source position $\sBif_i$ to the target position $\tBif_j$.  
Since the local stiffness matrices of parenchyma and vascular elements depend on the actual displacement vector $\bu$, the problem is non-linear, and the final equilibrium must be computed iteratively. A damped Newton-Raphson method is used: in each iteration $k$, the update $\Delta\bu^{(k+1)}$ of nodal positions is computed by solving a system of linear equations
\begin{equation}
\left[\tau\bI + \hat{\bK}_\pi(\bu^{(k)})\right]\Delta\bu^{(k+1)} = -\bg_\pi(\bu^{(k)})
\label{eq:NR}
\end{equation}
where $\hat{\bK}_\pi$ is the global system matrix after imposition of non-homogeneous Dirichlet boundary conditions corresponding to the hypothesis $\pi$, $\tau$ is a damping parameter, $\bI$ identity matrix having the identical size as $\hat{\bK}$ and the vector $\bg$ of the right side gathers the internal elastic forces and prescribed displacements given by the hypothesis $\pi$. The displacement vector is updated in each step of the method as $\bu^{(k+1)} = \bu^{(k)} + \Delta\bu^{(k+1)}$. the equilibrium displacement obtained for hypothesis $\pi$~\cite{Peterlik2018}.
Since the aim is to minimize the time needed for the computation of the transformation, \CoB{ preconditioning is used}: In the first iteration of the Newton-Raphson method, the Eq.~\ref{eq:NR} is solved with an algorithm based on the sparse Cholesky decomposition\footnote{\url{https://www.pardiso-project.org/}}. In the following iterations, preconditioned conjugate gradients are used to compute the update $\Delta\bu$ employing the decomposition constructed in the first iteration.

Besides the elastic transformation, the FE model is employed to obtain the compliance in the \textit{free source bifurcations $(\sBfree)$} which is used in both the \vesselGM{} and \adaptGM{} algorithms. 
According to the mathematical definition, compliance is defined as the inverse of the stiffness matrix. It is evaluated in an arbitrary node $n$ of the FE mesh where it is given by a 3$\times$3 symmetric tensor $\bC^n$ extracted from the global compliance matrix. Eigenvectors of the compliance tensor define the principal axes of an ellipsoid and the eigenvalues determine its scale along each axis. From the mechanical point of view, this ellipsoid characterizes the flexibility of $n$, \ie, it is proportional to the volume to which the node can be displaced under constant unit force applied to the node $n$ in an arbitrary direction.
The source bifurcations coincide with a mesh node $i$ hence its compliance $\bC^{i}_\pi$ corresponds to the block extracted from $\hat{\bK}^{-1}_\pi$ at the position indexed by $i$. Since the proposed method is adapted to large deformations, it is necessary that the compliance $\bC^{i}_\pi$ is computed using the stiffness matrix $\hat{\bK}_\pi$ computed using the equilibrium displacement vector $\bu_\pi$. Therefore, the elastic transformation corresponding to hypothesis $\pi$ is computed before the $\bC^{i}_\pi$ is obtained for each \textit{free source bifurcation}.

The compliance has been used in other applications to produce structures having desirable physical properties \cite{MartinezStructShape2015}. Whereas, \CoB{in the following matching algorithms} the compliance defines an improved metric for the generation hypotheses.

\subsection{Vessels Compliance Graph Matching (\vesselGM{}) }
\label{ss:meth:compMatch}

\begin{figure}[H]
	\centering
	\adjincludegraphics[width=1.0\linewidth]{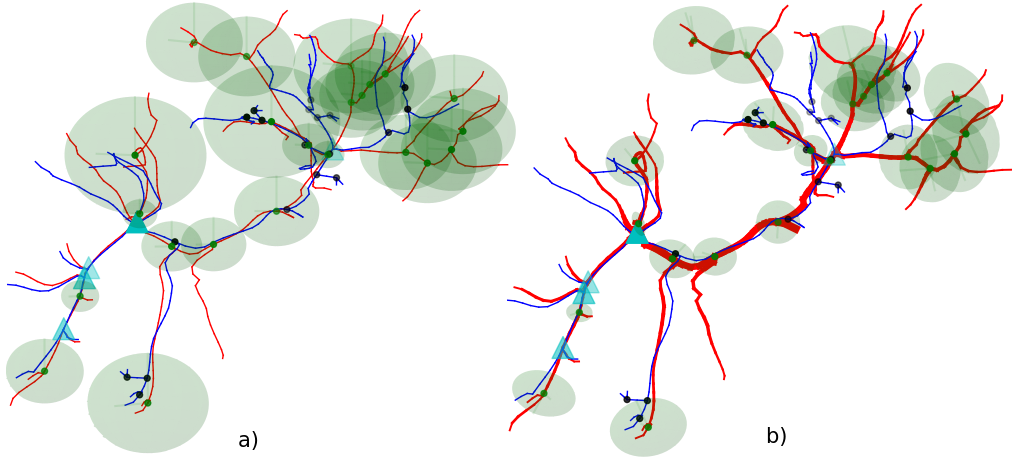}	
	\caption{ Using the initially matched bifurcations (triangles), the bounded regions (spheres) are computed in the free bifurcations (green dots) of the source graph (in red). The free bifurcations (black dots) of the target graph (in blue) inside the bounded regions define the next matching candidates. The bounded regions are defined (a) in \initializedFEM\ with the GPR covariance and (b) in \vesselGM\ with the compliance of the varying radii vessels. In the later case, the stiff thick vessels have smaller compliance. This, along with the tensors shape and orientation reduce the matching search space.}  
	\label{fig:CovarianceComplianceMatch}
\end{figure}

\CoB{\initializedFEM{}~\cite{GarciaGuevara2018} uses the GPR covariance to define the hypotheses search space and is able to find correct bifurcation matches even when large nonlinear deformations occur.} 
However, the covariance produces large bounded regions, shown in Fig.~\ref{fig:CovarianceComplianceMatch}.a, and large computation time. This makes the algorithm incompatible with intra-operative deployment. 

To overcome this crucial limitation of \initializedFEM{}, the first improvement proposed is \vesselGM{} which is described in Algorithm~\ref{alg:compGM}. This algorithm matches a source graph $\grph^S=(\vec{X}^S, \vec{E}^S)$ to a target graph $\grph^T=(\vec{X}^T, \vec{E}^T)$. 
Where $\vec{X}^S=\{\mathbf{s}_n$, $n = 1 ... N_S\}$ and $\vec{X}^T=\{\mathbf{t}_n$, $n = 1 ... N_T\}$ are the sets of source and target nodes (tree bifurcations). 
Similarly, $\vec{E}^S$  and  $\vec{E}^T$ are the sets of source and target edges.

The algorithm is initialized with an incomplete hypothesis $(\pi_{iGPR})$ which is a set of matched bifurcations $\sBif_i \leftrightarrow \tBif_j$ from the source and target graphs (line 1).
In practice, $\pi_{iGPR}$ does not include matches implying large deformations and is quickly obtained with \improvedGPR{}. 
From this initialization, the algorithm generates a set of hypotheses recursively.
\CoB{The current hypothesis $(\pi_t)$ defines the free source and target bifurcations sets as $\sBfree=\{\mathbf{s}_n \colon \vec{X}^S \notin \pi_t\}$ and $\tBfree=\{\mathbf{t}_n \colon \vec{X}^T \notin \pi_t\}$, respectively. 
Up to here,  \vesselGM{} is similar to \initializedFEM{}.}

\begin{algorithm}[H]
	\caption{Recursive vessels compliance graph matching (\vesselGM{}) $\grph^S,\grph^T$}
	\label{alg:compGM}
	\begin{algorithmic}[1]
		\Let{$\pi_{iGPR}$}{$\{s_1 \leftrightarrow t_1, ... , s_K \leftrightarrow t_K\}$}
		\Comment{\improvedGPR{} matching initialization}		
		\Function{recursiveGraphMatching}{$\pi_{t}$}
		\State{$\mFEM, \comp $ = \textbf{simulationFEM}($\pi_{t},\grph^S,\grph^T$)} 
		\State $\matchFew$ = \Call{\textbf{FindCandidates}}{$\mFEM, \comp$}	
		\State{$S_{\pi_t}$ = QualityScore($\mFEM$ )} 
		\If{$|\matchFew| \ne 0 $} 
		\For{$\PotCand_{i^*}$ in RandomPermutation($\matchFew$) } 		
		\For{$\tBif_{j^*}$ in RandomPermutation($\PotCand_{i^*}$) } 
		\Let{$\pi_{t+1}$}{$\pi_{t} \cup \{ \sBif_{i^*} \leftrightarrow \tBif_{j^*} \}$}
		\State\Call{RecursiveGraphMatching}{$\pi_{t+1}$}
		\EndFor		
		\EndFor		
		\EndIf		\EndFunction
		\State $\pi^*$ = $\argmax{ \{ S_{\pi_{iGPR}}, ..., S_{\pi_T} \} }$ 
		
		\Statex		
		\Function{\textbf{FindCandidates}}{$\mFEM, \comp$}
		\For{$\sBif_{i}$ in $\sBfree$}
		\State{$ \mathcal{B}_{i} = \{ \forall \ \tBif_j \in \tBfree \colon \mid M_{comp}^2 (\mFEM(\sBif_i), \ \tBif_j) < MC_{TH} \lor |\mFEM(\sBif_i)- \tBif_j| < E_{TH}  \} $}
		\State{$\PotCand_i$ = $\{ \tBif_j \colon \tBif_j \in |\tMGEO-\sMGEO|<(\thGEO)(\sMGEO) \land \tBif_j \in \mathcal{B}_i \} $ }
		\EndFor
		\State{$\matchFew$ = $\argmin_i \{| \PotCand_i | \} $ for $|\PotCand_i| \ne 0$ } 
		\EndFunction		
	\end{algorithmic}
\end{algorithm}

\CoB{\vesselGM{}'s contribution starts by using $\pi_t$ with} the FE simulation to compute the transformation $(\mFEM)$ and the compliance $(\comp)$ at $\sBfree$ (line 3). 
Here, the $\mFEM$ and $\comp$ replace, respectively, the GPR mean and covariance, removing completely this dependency. 

To find the next matching candidates $\matchFew$ (line 5) the Mahalanobis distance
\begin{equation}\begin{split}
\label{eq:MahalanobisCompliance}
M_{comp}^2 = (\mFEM(\sBif_i) - \tBif_j)^T (\bC^{i}_{\pi_{t}})^{-1} (\mFEM(\sBif_i) - \tBif_j )
\end{split}\end{equation}
\CoB{computed with the compliance $\comp$ is used. Here, $M_{comp}^2$ is equivalent to the virtual work needed to add a candidate match. 
The biomechanical simulation is important because $\mFEM$ transforms the source bifurcations close to the target ones and the beam model correctly simulates the deformation along the graph edges.
In addition, the compliance tensor filters candidates that require high energy to match.
Especially,  the beam model introduces additional stiffness and anisotropy that leads to smaller compliance in thicker vessels.
This reduces the bounded region of these rigid vessels while keeping higher compliance and bounded regions on thin flexible vessels. 
These compliance-based bounded regions are shown in Fig.~\ref{fig:CovarianceComplianceMatch}.b. }

For each free bifurcation in $\sBfree$, the bounded region candidates $(\mathcal{B}_{i})$ are the target free bifurcations $(\tBfree)$ within Mahalanobis (\thMC) or strict euclidean ($E_{TH}$) distance thresholds (line 18). 

\CoB{From here, the algorithm is again similar to \initializedFEM{}}. In line 19, $\tMGEO$ are the target geodesic distances from the already established correspondences $(\pi_{t})$ to a new match  $\tBif_{j}$. Similarly, $\sMGEO$ denotes the source geodesic distances. 
Thus, the potential candidates $(\PotCand_i)$ are the bifurcations for which target and source geodesic distances are similar.  
Then, the matching candidates $\matchFew$ found  are the free bifurcations $\sBfree$ with the lowest number of potential candidates (line 21).  

\CoB{The algorithm continues (line 5) with the computation of the current hypothesis's quality score~\cite{Serradell2015NonRigid}. Then, in the nested loops the hypotheses are generated. First, every free bifurcation in the random permutation of the matching candidates $(\matchFew)$ is explored.
Then, every target candidate ($\tBif_{j^*}$) with its associated free source bifurcation from the random permutation of $\PotCand_{i^*}$ creates a new hypothesis ($\pi_{t+1}$).}
The new hypothesis is used to recursively call the matching method until no more matches are found. 
Finally, the best quality score hypothesis is selected from all the explored ones.


\subsubsection{Setting the Mahalanobis compliance threshold}
\label{ss:meth:compMatchSet}

\begin{figure}[H]
	\centering
	\adjincludegraphics[width=1.0\linewidth]{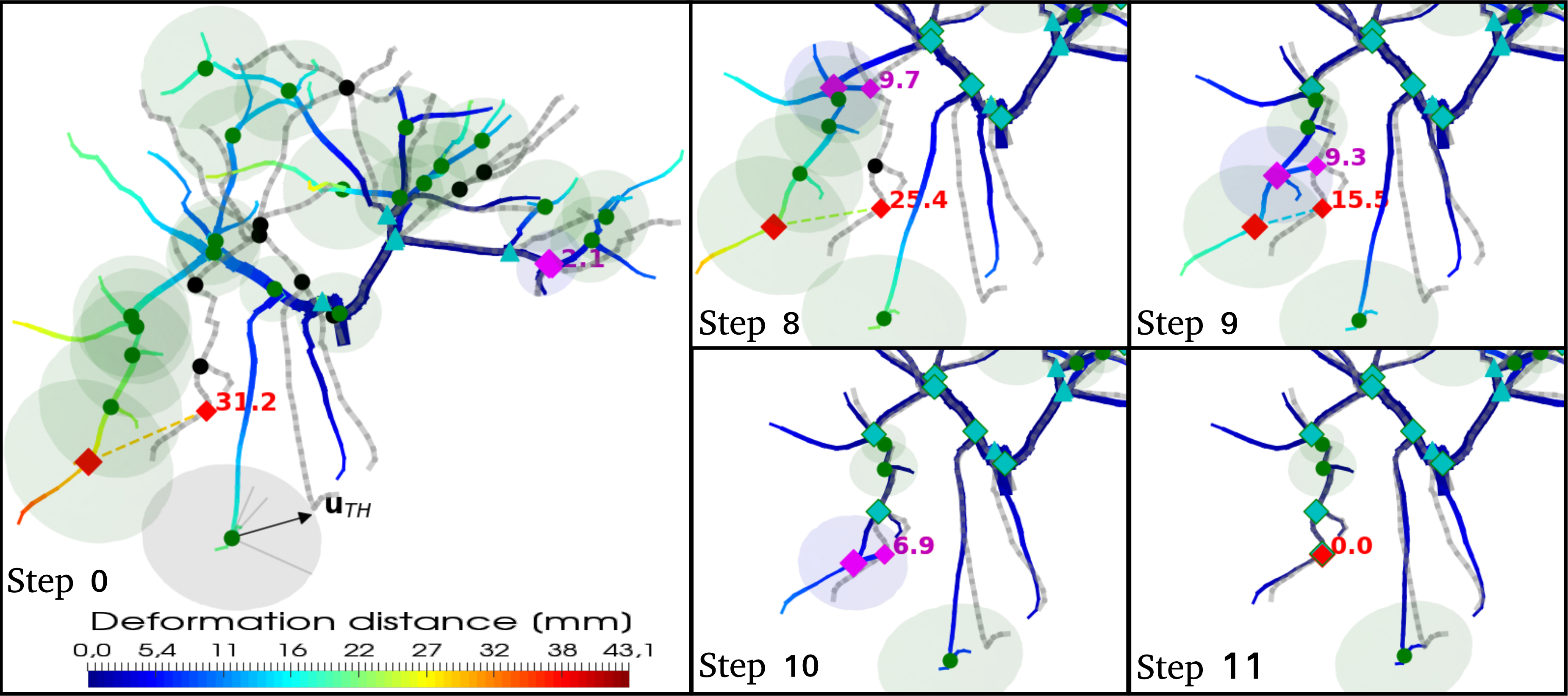}
	\caption{ 
			The source and target graphs are color-coded with the registration error magnitude and in gray, respectively. 
			In the first recursive step, the $\pi_{iGPR}$ initialization (cyan triangles matches) and the FE simulation determine the most flexible bifurcation (inside the gray bounded region). This bifurcation's compliance $\bC_{max}$ and the maximum incremental displacement $\bu_{TH}$ define the $MC_{TH}$ threshold. 
			Then, every recursive step adds a new match (connected diamonds in pink with its displacement magnitude $|\mathcal{T}_{\pi_{t}}(\sBif_{i^*})-\tBif_{j^*}|$) to the current hypothesis.	
			This incremental hypotheses generation progressively reduces the source graph's registration error, including the initial maximum deformation $(|\mathcal{T}_{\pi_{iGPR}}(\sBifMax)-\tBifMax|)$ of the connected diamonds match in red. } 		
	\label{fig:VCGMincrementalHypo}
\end{figure}

\JG{In \initializedFEM{}, the covariance Mahalanobis threshold ($M_{TH}$) specifies a level of confidence given by the GPR covariance. However, setting an optimal $M_{TH}$ remains highly dependent on each dataset (e.g. level of deformation, noise). }
In \vesselGM{}, the \thMC{} threshold represents an upper bound of the work needed to match bifurcations. 	
\JG{The optimal \thMC{} depends on the FE model, deformation magnitude, initialization and incremental hypotheses generation (depicted in Fig. \ref{fig:VCGMincrementalHypo}).} 
	
\CoB{\JG{To simplify the optimal setting of \thMC{},} it is defined as a function of the maximum incremental displacement's magnitude $(U_{TH}=|\bu_{TH}|)$:}
	
\begin{equation}\begin{split}
\label{eq:MahalanobisComplianceThreshold}
MC_{TH} = f(U_{TH})=(\bu_{TH}^T (\bC_{max})^{-1} \bu_{TH})^{1/2},
\end{split}\end{equation}

where $\bC_{max}$ is obtained using the FE simulation at the initialized stage ($\pi_{iGPR}$) by selecting the bifurcation whose compliance ellipsoid volume is the maximum, i.e. the most "flexible" bifurcation. 
\CoB{Since the free bifurcations can move in any direction, the maximum incremental displacement $\bu_{TH}=(U_{TH}/\sqrt{3})(\hat{e}_1+\hat{e}_2+\hat{e}_3)$ is assumed isotropic with respect to  the eigenvectors of $\bC_{max}$. }

\CoB{
The optimal $U_{TH}$ remains dependent on the incremental hypotheses generation and the deformation magnitude. Still, it is assumed that $U_{TH}$ is smaller than the initial maximum match's deformation $(|\mathcal{T}_{\pi_{iGPR}}(\sBifMax)-\tBifMax|)$ and bigger than the new matches' displacements $(|\mathcal{T}_{\pi_{t}}(\sBif_{i^*})-\tBif_{j^*}|)$.  
In practice, setting $U_{TH}=14\ mm$ (about 40\% of the largest bifurcation's deformation) allowed to match successfully several experiments, which have a wide range of initial maximum displacements. 
}

Since, $MC_{TH}$ is larger than required in some recursive steps of the algorithm, 
using a constant $U_{TH}$ unnecessarily increases the search space. 
\CoB{\adaptGM{} alleviates this issue.} 

\subsection{Matching based on adaptive Mahalanobis distance (\adaptGM{})}
\label{ss:meth:adaptMatch}

In \vesselGM{},  only the bifurcations that require less than a given amount of work, bounded by the constant compliance Mahalanobis threshold $(MC_{TH})$, \CoB{are matched}. 
Although the compliance tensor filters some incorrect matches, this constant upper bound unnecessarily increases the search space. 
This is because the incremental matching (in Fig. \ref{fig:VCGMincrementalHypo}) does not always need the constant upper bound at every recursive step.

Instead of setting a constant threshold, \adaptGM{} uses the range $[MC_{Low}, MC_{High}]$. 
\CoB{As presented in the Algorithm \ref{alg:adaptGM}, the} rigid-to-soft approach starts by adding the bifurcation matches that require the least amount of work ($MC_{Low}$) and when no more matches are found, instead of exploring other alternative hypotheses, the work bound is gradually increased. 
Hence, the matches that require more work are gradually added until a maximum allowed work ($MC_{High}$) is reached (lines 6 to 9).

In most cases, the rigid-to-soft strategy finds an appropriate set of correspondences before exploring an alternative matching path. 
Thus, when the exploration of the first matching path is finished, $MC_{High}$ is reduced to save time (line 16).

\CoB{
$MC_{High}$ can be set higher than the optimal value (overestimated) to guarantee that it covers a wide range of deformation, scale, or incremental exploration dependencies. Thanks to the rigid-to-soft approach, an overestimated $MC_{High}$ produces only a small increase in computation time.
}

\JG{
When there are outlier  matches that require less work to be matched than the correct matches, $MC_{Low}$ is a critical parameter. In this specific case, $MC_{Low}$ should be large enough to include the correct matches. 
}

\todo[inline]{plutot: however, these outliers are mostly filtered  out by the geodesic constraint; }
\todo[inline]{This is mostly about the case when geodesic constraint is not able to filter outliers, as shown in experiment 3.1.2 with 50\% noise and explained at the beginning of the discussion }


\begin{algorithm}[H]
	\caption{Recursive Adaptive compliance FEM matching $\grph^S,\grph^T$}
	\label{alg:adaptGM}
	\begin{algorithmic}[1]
		\Let{$\pi_{iGPR}$}{$\{\sBif_1 \leftrightarrow \tBif_1, ... , \sBif_K \leftrightarrow \tBif_K\}$}
		\Comment{\textbf{\improvedGPR{} matching initialization}}		
		\Function{recursiveAdaptiveGraphMatching}{$\pi_{t}$}
		\State{$\mFEM, \comp $ = \textbf{simulationFEM}($\pi_{t},\grph^S,\grph^T$)} 
		\State{$S_{\pi_t}$ = QualityScore($\mFEM$ )} 
		\State {$\MCpt=MC_{Low}$}
		\While{$\MCpt<MC_{High} \wedge |\matchFew| = 0 $} 
		\State $\matchFew$ = \Call{FindCandidates}{$\mFEM, \comp$}		
		\State{$\MCpt=\MCpt+(MC_{High}-MC_{Low})/6$} 
		\EndWhile			
		\If{$|\matchFew| \ne 0 $} 
		\For{$\PotCand_{i^*}$ in RandomPermutation($\matchFew$) } 		
		\For{$\tBif_{j^*}$ in RandomPermutation($\PotCand_{i^*}$) } 
		\Let{$\pi_{t+1}$}{$\pi_{t} \cup \{ \sBif_{i^*} \leftrightarrow \tBif_{j^*} \}$}
		\State\Call{RecursiveGraphMatching}{$\pi_{t+1}$}
		\EndFor
		\State {$MC_{High} = (MC_{High} + \MCpt)/2$}	
		\EndFor
		\EndIf		\EndFunction
		\State $\pi^*$ = $\argmax{ \{ S_{\pi_{iGPR}}, ..., S_{\pi_T} \} }$ 
		\Statex		
	\end{algorithmic}
\end{algorithm}


\subsubsection{Setting Mahalanobis compliance threshold range for \adaptGM{} method}
\label{ss:meth:adaptMatchSet}

\adaptGM{} also uses Eq. \ref{eq:MahalanobisComplianceThreshold} to define the range $[MC_{Low}, MC_{High}]$ as a function of $[U_{Low}, U_{High}]$. 
\CoB{
The initialization, FE model and $U_{Low}$ help to define a correct $MC_{Low}$.
Therefore, even in the specific case of outliers that are not geodesic-filtered, $U_{Low}$ is a less critical parameter.} 
\JG{$U_{High}$ is similar to $U_{TH}$, they indirectly depend on deformation and scale because of the incremental hypotheses generation. The advantage of $U_{High}$ is that it is overestimated to cover a wide range of cases without a high increase in computation time.} 
\todo[inline]{ pourquoi the incrmental exploration cases? }
\todo[inline]{ Imagine the case in Fig. 3 that the bifurcations match in step 9 were missing, then the incremental matching would change and then the Uth should be 15.5 to match the last bifurcation.}

\todo[inline]{In Section 2.3.1, it is stated that Ulow of 9mm can produce successful match but the reviewer thinks this is dependent on the scale of the vasculature. The authors are therefore suggested to rephrase it to clarify this is a case-specific (unless the reviewer is misunderstanding it). }
\smallskip%
\section{Results}
\label{s:results}
\smallskip%

This section \CoB{presents} the case of augmenting intra-operative Cone Beam Computed Tomography (CBCT) images with preoperative computed tomography angiography  (CTA) data. The expected benefit is an improved visualization of the patient's tumor(s), vascular system and other internal relevant structures.
CBCT is an imaging modality that is more available in the operating room than CTA or MRI.
However, contrast to noise ratio in CBCT is about half than that in CTA. CBCT suffers motion artifacts, beam hardening, partial volume and ring effects \cite{Tacher2015coneBeamCT}. 
Thus, CBCT cannot image certain lesions nor complete anatomy. These deficiencies are compensated if preoperative data augment the intra-operative image.
The fusion approach proposed is based on the matching of pre- and intra-operative vascular trees according to the methods presented above and evaluated on both synthetic and real data. All the results and computation times were obtained with a regular desktop computer (4GHz eight-core, 16 GB RAM).

\smallskip%
\subsection{Experiments on synthetic data}
\label{s:res:synth}
\begin{figure}[H]
	\centering
	\adjincludegraphics[width=1.0\linewidth, trim={0  0 0 0},clip]{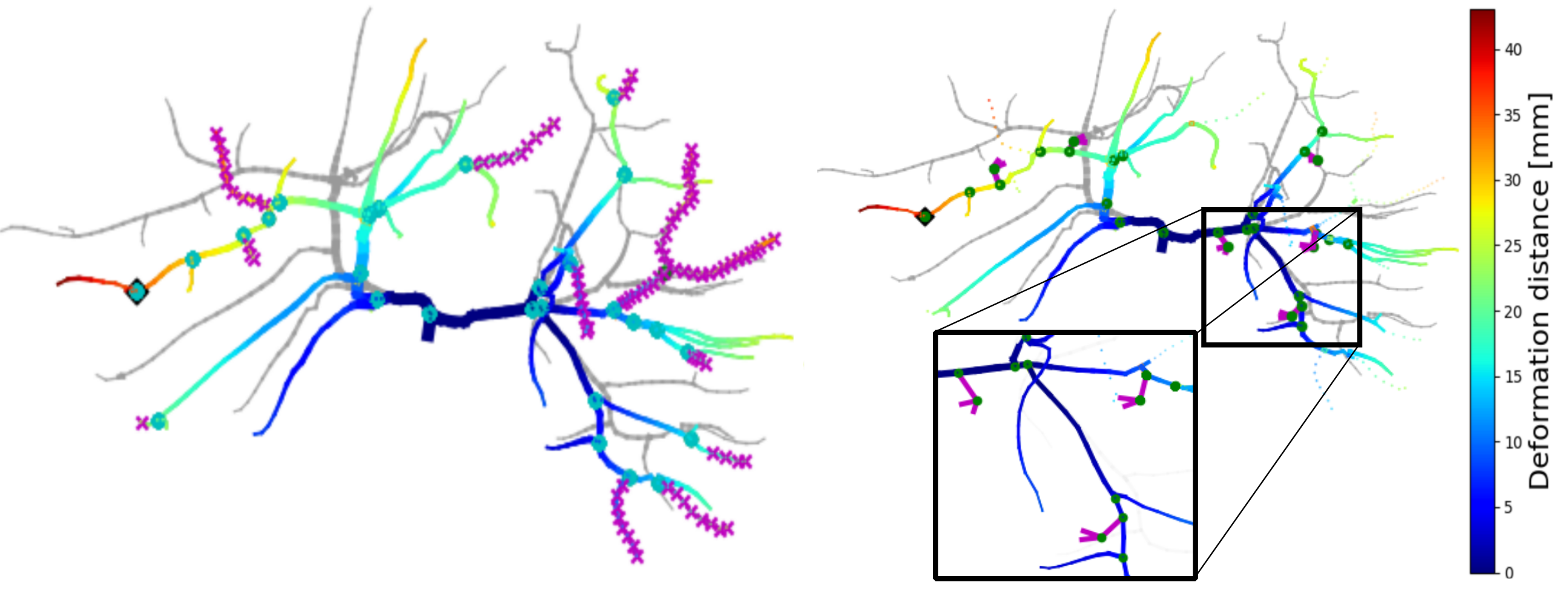}
	\caption{The initial vascular graph (in gray) and the synthetic target graph (color-coded on deformation amplitude) are used for evaluation of the method. 
	On the left, the magenta crosses represent the removed branches (40\% of the initial graph). On the right, noisy branches (in magenta) are added to the deformed and reduced target graphs.}
	\label{fig:SynthDeformation}
\end{figure}

To evaluate the methods, synthetic (target) graphs that resemble CBCT were generated from a real vascular graph which was segmented from a CTA image. 
First, the original CTA graph was deformed using a realistic hyperelastic FE simulation that simulates the effect of the pneumoperitoneum on the liver~\cite{GarciaGuevara2018}. 
\FIX{It is important to note that the FEM model used for matching is linear which makes the computation faster and uses a 1.5\,kPa Young Modulus which is different from the ground truth simulations.}
From this deformed CTA graph, leaf vessels were iteratively removed until only 60\% remained, as a way to mimic the partial graph usually segmented from CBCT images. Then, noisy bifurcations were added (50\% or 80\% of the bifurcations remaining from the previous step).
An example of the synthetic data generated is shown in Fig.\,~\ref{fig:SynthDeformation}.
The original CTA graph and the deformed, reduced, and noisy target graphs were used to evaluate and compare the matching methods presented in this article.  
The target registration error (TRE) is computed in the complete original graph, including the 40\% of vessels that were removed (therefore the TRE cannot reach zero in these experiments).
\subsubsection{Accuracy and search space size with increasing Mahalanobis thresholds} 
\label{ss:res:synth:IncMah}

The first experiment evaluates the search space size and accuracy as a function of the bound region used.
Matching was repeated 10 times per method with increasing Mahalanobis thresholds. 
\CoB{\thMC{}, $MC_{Low}$ and $MC_{High}$ were set without using Eq. \ref{eq:MahalanobisComplianceThreshold}.} \todo[inline]{what do you mean by directly set?}\todo[inline]{Kind of explain in discussion}
These different matchings were only used to study the methods' Mahalanobis threshold sensitivity.
Increasing thresholds were tested in one synthetic dataset (28 source and 16 target bifurcations) with 4 different initializations (named $E_n$ with $n \in [1,4]$). 
Initialization $E_3$ has 6 matches while the other initializations have 5 matches, that is why $E_3$ explores fewer hypotheses. The three evaluated methods use the same realistic Euclidean ($E_{TH}=4 \ mm$) and geodesic thresholds ($G_{TH}=20\%$).

\begin{figure}[H]
	\centering
	\includegraphics[width=1.0\linewidth, height=4cm]{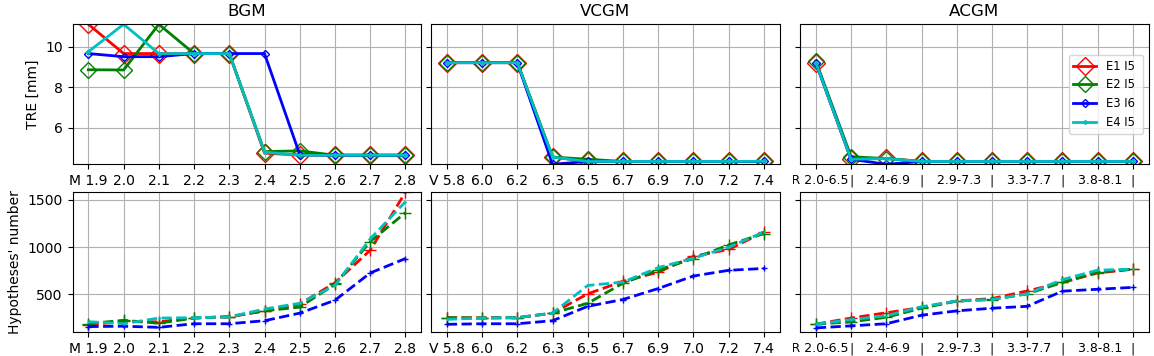}
	\caption{Matching accuracy and search space size with increasing Mahalanobis thresholds in one synthetic dataset without noise. 
	\thMC{} and $MC_{Low}$-$MC_{High}$ were set without using Eq. \ref{eq:MahalanobisComplianceThreshold} and are specified in the horizontal axis.		
	The top row shows the TRE for each method  with four different matching initializations ($E_n$) and the initial number of matches ($I_k$). 
	Similarly, the bottom row shows the respective required number of hypotheses explored. 		
	The search space is better pruned with the  \vesselGM{} and \adaptGM{} methods while the exploration highly increases with higher Mahalanobis thresholds in \initializedFEM{}. }
	\label{fig:convergenceAllExperimentsMatching}
\end{figure}

For every method, as shown in Fig.\,~\ref{fig:convergenceAllExperimentsMatching}, the TRE remains high when a small Mahalanobis threshold is used, but  decreases as the threshold increases. 
However, if a high threshold is used the number of explored hypotheses and search time increases 
without a significant improvement of the TRE.   
Nevertheless, \initializedFEM{}'s search space increase is steeper than for \vesselGM{} or \adaptGM{}. 
\JG{\adaptGM{} does the best pruning of the search space and has the best TRE results with less dependency on the Mahalanobis threshold used.}

 %
\subsubsection{Accuracy and search space size at the optimal Mahalanobis threshold}
\label{ss:res:synth:optMah}

\begin{figure}[H]
	\captionsetup{aboveskip=-1pt,belowskip=-15pt}
	\centering
		\adjincludegraphics[width=1.0\linewidth, trim={0 0 0 0 },clip]{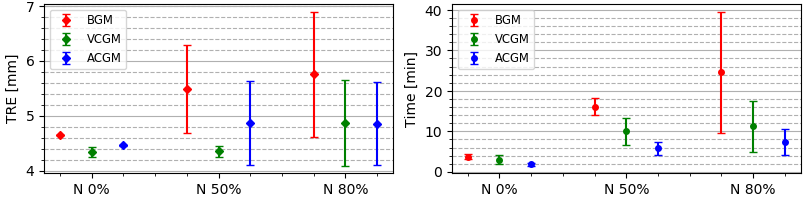}
	\caption{
		TRE (left) and matching time (right) statistics obtained on 4 different initializations using the optimal Mahalanobis threshold (where \thMC{} and $MC_{Low}$-$MC_{High}$ were set without using Eq. \ref{eq:MahalanobisComplianceThreshold}). 
		Non-noisy graphs as well as graphs with bifurcations noise (50\% and 80\%) are considered in the study.
		The \vesselGM{} and \adaptGM{} methods have better TRE than \initializedFEM{} . When noise is added, these methods require less matching time than \initializedFEM{}. 
		\adaptGM{} is the fastest method overall.}
	\label{fig:DeformationAndStats}
\end{figure}

Using the previous experiment, every method is also compared at its optimal performance, i.e. the smallest Mahalanobis threshold at which the TRE reaches a minimum value.  
The mean and standard deviation for 4 different initializations at the optimal threshold are plotted in Fig.\,~\ref{fig:DeformationAndStats}. 
Similar experiments with added noisy bifurcations (50\% and 80\%) are also plotted.  
Without noise, the \vesselGM{} and \adaptGM{} methods are faster than \initializedFEM{}.
When noise is added, the exploration space is highly increased with \initializedFEM{}.
This case requires up to 40 minutes \JG{to do the matching} and sometimes fails to find the correct match, increasing the mean TRE up to 5.8 mm. 
\JG{Differently, the compliance methods require, on average, less than 12 minutes. 
\adaptGM{} is the fastest with less than 8 minutes on average, and has 4.8 mm mean TRE.} 

\subsubsection{Synthetic deformations using the same matching parameters}
\label{ss:res:synth:samePar}

Ten different deformations were generated using different pressure, Young's modulus and gravity orientation (to simulate subject in supine or flank position). Also, a craniocaudal force simulated different respiratory phases. These deformations are summarized in Table~\ref{Table:SynthDefStats} with statistics of the bifurcations' displacements. 
As in the previous experiments, the graphs have 40\% branch removal and two levels of noisy bifurcations added. 
\begin{table}[H] 
	\centering 
	\caption{The graph bifurcations displacements statistics of the ten synthetic deformations.} 
	\label{Table:SynthDefStats} 
	\begin{adjustbox}{max width=\textwidth}
	\begin{tabular}{|m{24mm}|c|cccccccccc|}
		\toprule
		\multicolumn{2}{|c|}{Deformation dataset} & D1 & D2 & D3 & D4 & D5 & D6 & D7 & D8 & D9 & D10 \\
		\midrule
		\multirow{3}{24mm}{Bifurcations displacements [mm] }
		& $\mu$ & 10.2 & 12.1 & 13.9 & 13.5 & 8.6 & 11.6 & 14.8 & 12.2 & 9.0& 11.0 \\
		&$\sigma$ & 5.9 & 6.8 & 8.5 & 8.1 & 6.9 & 7.3 & 9.4 & 6.9 & 5.0& 6.8 \\
		&max& 21.2 & 24.7 & 31.3 & 29.2 & 26.7 & 25.1 & 35.6 & 25.8 & 17.7& 23.9 \\
		\bottomrule
	\end{tabular}
\end{adjustbox}
\end{table}

\JG{Each method uses constant parameters to match all different synthetic datasets deformations.
With  \initializedFEM{}, the optimal $M_{TH}$ depends on the deformation magnitude and the noise, therefore being potentially difficult to set in clinical scenarios where no information of the deformation is \textit{a priori} known. For this reason, optimal threshold $(M_{TH}=2.6)$ found from the previous experiments \CoB{is used}. This threshold, using a cumulative chi-squared distribution, represents a 97\% confidence region.}


For the \vesselGM{} and \adaptGM{} methods, 
the parameters selection is simplified since the \thMC{} takes into account the FE model and the initialization. 
Thus, the only parameter to set are the maximum incremental displacement magnitudes. \JG{Given that the compliance methods are not too sensitive to these parameters, approximate values are sufficient. \vesselGM{} \CoB{uses} $U_{TH}=14\ mm$, whereas \adaptGM{} uses $U_{Low}=9\ mm$ and $U_{High}=15\ mm$ to define $[MC_{Low}, MC_{High}]$. }
%
%
%

\begin{figure}[H]
	\captionsetup[subfigure]{aboveskip=-4pt,belowskip=4pt}
	\centering
	\adjincludegraphics[width=1.0\linewidth, trim={0  0 0 0},clip]{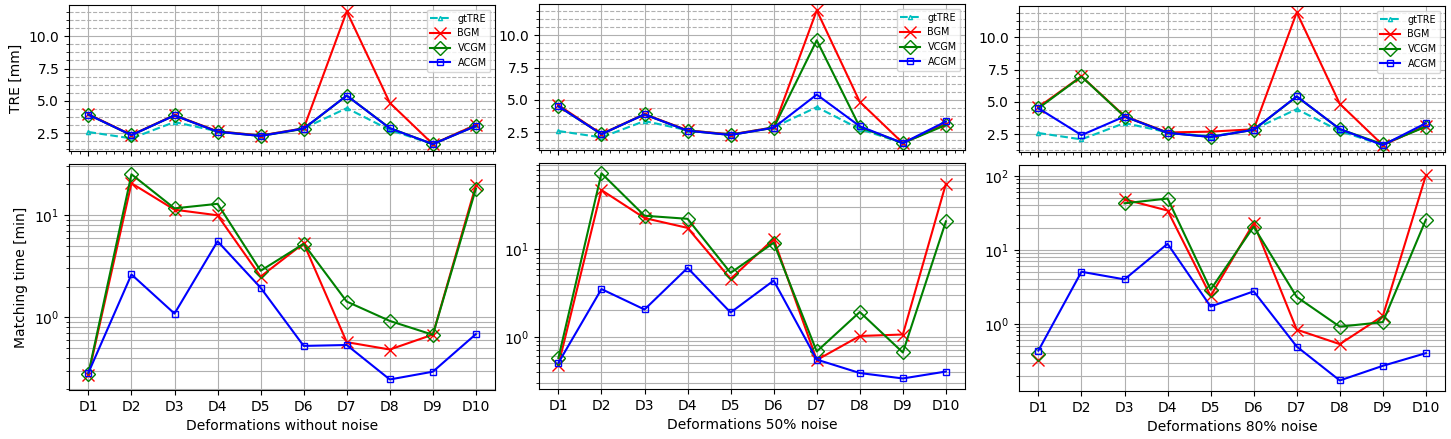}
	\caption{
	The TRE and the matching time (in logarithmic scale) are evaluated for the three matching methods on ten different deformations using constant covariance Mahalanobis $(MC_{TH})$ and maximum  exploration displacement thresholds $(U_{TH}$, $U_{Low}$ and $U_{High})$.
	Each plot has the results of graphs with bifurcations noise (50\% and 80\%) and without it. The reference TRE (2.76$\pm$0.73 (4.42) mm) is computed with the ground truth bifurcations matches and shown in cyan.
	} 
	\label{fig:tenSynthDefTHMC}
\end{figure}

\todo[inline]{Figure 7, ACGM method results are farther than the other two methods (VCGM and BGM methods). How do you explain that ACMG is the best method? Which was the reference to obtain such conclusions? Authors should explain this in the paragraph "The ACGM method is the fastest and finds all the correct matches even the noise. The other methods fail to find the correct match in some cases (D2, D7 and D8)."}
The Fig.\,~\ref{fig:tenSynthDefTHMC} presents the matching time 
\CoB{and the TRE measured with synthetic ground truth data of each experiment. 
\adaptGM{} has the same or better (in some cases of deformations D2, D7 and D8) TRE and  is faster than the other methods. }

\todo[inline]{mo: in legenf of fig.7 I suppose you take minimum  values for  (Thmin)?}

Table\,~\ref{Table:SynthDefStatsTHMC} summarizes the results of the ten deformations at each level of noise. 
Without noise, the average matching time of \adaptGM{} is 1.4 minutes, while it is more than 7 minutes with the other methods.  
Given that \initializedFEM{} fails to find all the correct matchings, the average TRE is 0.8 mm worse than the other methods. 
Adding 50\% noise, the average matching time of \adaptGM{} only increases to two minutes, while it reaches 16 minutes with the other methods. 
With 80\% noise added, the trend is clear. 
The average matching time of \adaptGM{} only increases to 2.7 minutes, while it reaches 23 minutes with \initializedFEM{}. 
The noise highly affects \initializedFEM{}, with a TRE increasing to 4.5 mm because it did not complete all the experiments as some exhausted the computer RAM memory available.
\adaptGM{} maintains the best TRE result even with noise.

\CoB{The maximum incremental displacement range $[U_{Low}, U_{High}]$ simplifies the correct setting of the compliance Mahalanobis thresholds. 
This allowed to use constant parameters $(U_{Low}$ and $U_{High})$ to match correctly a diverse set of synthetic deformations, making \adaptGM{} very robust and efficient. 
}


\begin{table}[H] 
	\centering 
		\caption{Matching statistics  for ten synthetic deformations using constant covariance Mahalanobis $(MC_{TH})$ and maximum incremental displacement thresholds $(U_{Low}$ and  $U_{High})$.}
	\label{Table:SynthDefStatsTHMC} 
	\begin{adjustbox}{max width=\textwidth}
		\begin{tabular}{|c|c|c|c|c|c|c|}
			\toprule
			Noise  & \multicolumn{2}{|c|}{Without} & \multicolumn{2}{|c|}{50\%} & \multicolumn{2}{|c|}{80\%} \\
			Method  & TRE (max) [mm] & Time (max) [min] & TRE (max) [mm] & Time (max) [min] & TRE (max) [mm] & Time (max) [min] \\
			\midrule
			BGM & 3.9$\pm$ 2.8 (11.9) & 7.1$\pm$ 7.4 (20.4) & 4.0$\pm$ 2.8 (11.9) & 16.3$\pm$ 19.1 (55.5) & 4.5$\pm$ 2.9 (11.9) & 23.7$\pm$ 32.4 (102.4)\\
			VCGM & 3.1$\pm$ 1.0 (5.4) & 7.9$\pm$ 8.1 (24.9) & 3.5$\pm$ 2.2 (9.6) & 16.1$\pm$ 21.0 (73.1) & 3.6$\pm$ 1.5 (7.0) & 16.3$\pm$ 18.3 (49.7)\\
			ACGM & 3.1$\pm$ 1.0 (5.4) & 1.4$\pm$ 1.6 (5.5) & 3.2$\pm$ 1.1 (5.4) & 2.0$\pm$ 1.9 (6.1) & 3.2$\pm$ 1.1 (5.4) & 2.7$\pm$ 3.5 (12.1)\\
			\bottomrule
		\end{tabular}
	\end{adjustbox}
\end{table}

\subsection{Experiments on real data}
\label{s:res:RealData}

\CO{From the two porcine liver datasets (PA and PB),} each real dataset has one CTA image acquired preoperatively in supine position and one CBCT image acquired after pneumoperitoneum on flank position. 
These intra-operative conditions generated a large deformation. 
In both modalities the portal vein is visible, however, the CBCT image has fewer portal vessel branches visible than the CTA and several false branches were segmented due to noise (see Fig. \ref{fig:pipeline}, \CoB{the image intensity statistics in Table \ref{Table:VesselsStats}}). 
From the portal vein automatic segmentations, the source (CTA) and target (CBCT) graphs are extracted \cite{GarciaGuevara2018}.

\begin{figure}
	\captionsetup{aboveskip=-1pt,belowskip=4pt}
	\centerline{\adjincludegraphics[width=0.95\linewidth, trim={0 0 0 0},clip]{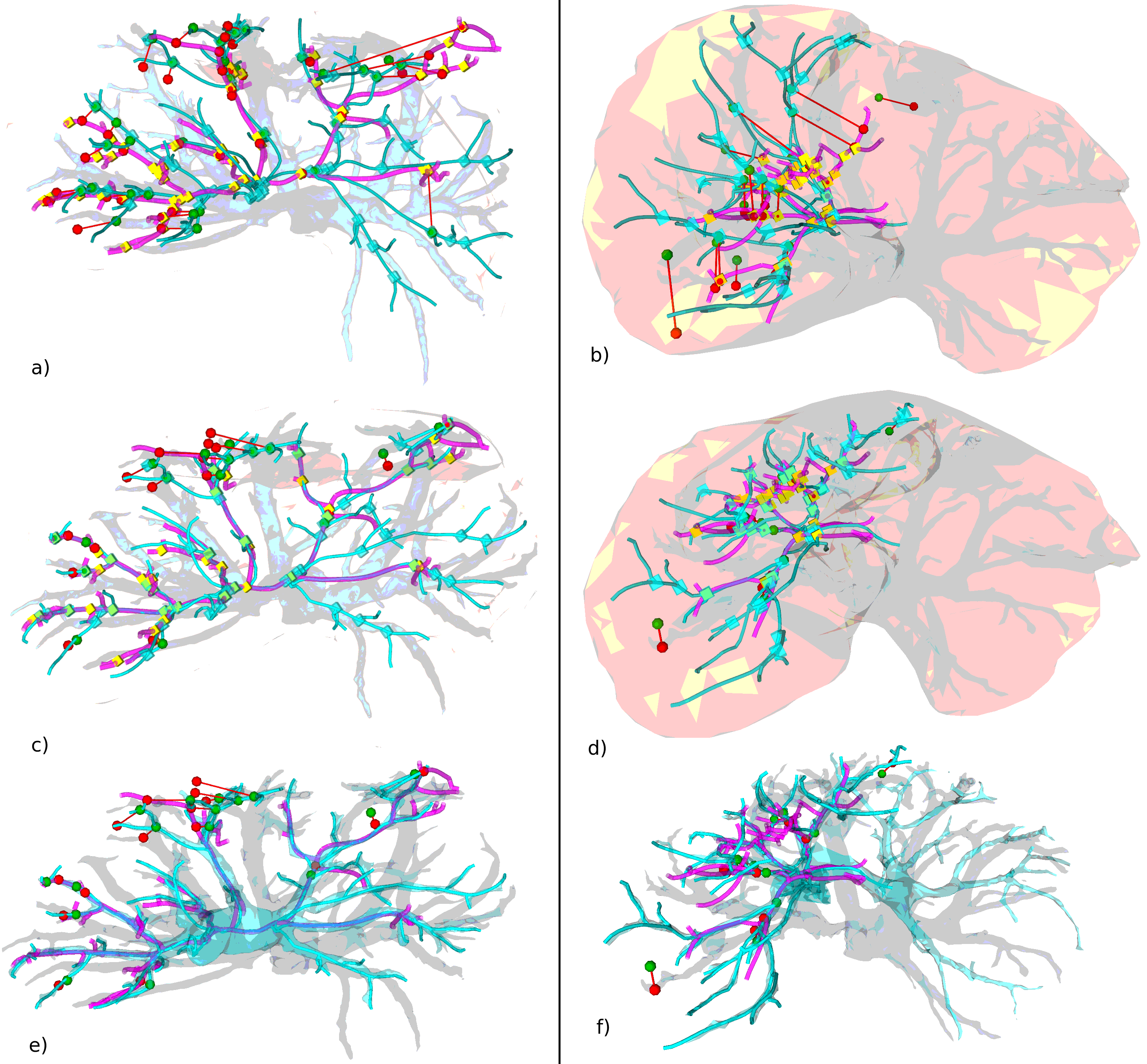}}
	\caption{
	The target CBCT (in pink) and source CTA (in cyan) portal vein graphs are rendered with tubular structures. 
	The graph nodes (bifurcations) are shown as cubic markers (in yellow for the target, cyan for the source and green for the matched).	
	The augmented hepatic vein, which was only visible in the CTA image, is in transparent blue behind the portal veins graphs.		
	In the left and right columns are the PA and PB porcine datasets, respectively. 
	The first row shows the 37 (for PA) and 22 (for PB) target evaluation landmarks (red spheres). It also shows the corresponding connected source landmarks (green spheres) and the liver structures rigidly aligned. These depict the large intra-operative non-linear deformation.
	The second and third row show the result of \adaptGM{} after the fine biomechanical matching \fineFEM{}. They also show the 16 (for PA) and 11 (for PB) evaluation landmarks with an error larger than 3 mm.
	The third row shows the transformed source portal vein used for matching (in cyan), instead of the augmented liver. 
	}
	\label{fig:matchFineLarge}
\end{figure}

The \CoB{portal veins} graphs are matched to register the CTA data onto the CBCT to augment it. For instance, the hepatic vein is not visible in the CBCT and fusing it from the CTA is clinically useful. 
Although the number of bifurcations (\CoB{in Table \ref{Table:VesselsStats}}) is similar in both modalities. Because of CBCT noise, there are several false bifurcations (most of the non-matched yellow cubes in the second row of Fig. \ref{fig:matchFineLarge}). 


\begin{table}[H] 
\centering 
\caption{The vessels characteristics of the porcine images.} 
\label{Table:VesselsStats} 
\begin{adjustbox}{max width=\textwidth}
\begin{tabular}{|c|c|cccc|} 
\toprule
\multicolumn{2}{|c|}{Porcine} & Number & Portal veins radii [mm]& Vessels intensity [HU] & Liver intensity [HU]\\
\multicolumn{2}{|c|}{dataset} & of nodes & [min, max] & $\mu$, min, max  & $\mu$, min, max  \\
\midrule
\multirow{2}{*}{PA}
& CT   & 60 & [1.6, 10.3] & 194.2$\pm$19.9  & 135.7$\pm$20.9  \\ 
& CBCT & 47 & [1.2, 8.8] & 176.7$\pm$39.2 & 103.8$\pm$39.8  \\ 
\midrule
\multirow{2}{*}{PB}
& CT   & 39 & [1.8, 7.5] & 189.1$\pm$23.4  &  117.0$\pm$25.2  \\ 
& CBCT & 36 & [1.1, 6.5] & 144.9$\pm$46.2  & 113.8$\pm$53.8   \\ 
\bottomrule
\end{tabular}
\end{adjustbox}
\end{table}


\CoB{The matching is evaluated with clear and unambiguous manually selected landmarks visible in both modalities. These landmarks include inserted tumors, distinctive vessel points and bifurcations (3, 15 and 19 for PA, while 6, 5 and 11 for PB, respectively). 
The registration error (RE) is computed using all these evaluation landmarks, while the matching only uses the bifurcations. Therefore, the RE cannot reach zero in these experiments. } 


The Table \ref{Table:RealDataTRE} presents the registration results for each dataset. 
First, using the matches obtained with the \improvedGPR{} initialization, \RW{the rigid RE quantifies the nonlinear deformation magnitude. Similarly, the first row of Fig. \ref{fig:matchFineLarge} depicts this deformation.} 
\JG{From \improvedGPR{} initialization, matching was repeated 10 times per method with increasing Mahalanobis thresholds to find the optimal. All these matchings are not needed during a normal registration and are only used to provide a meaningful comparison.}
\CoB{On average for the real datasets, \adaptGM{} has 2.34 mm and 0.58 mm better RE than \initializedFEM{} and \vesselGM{}, respectively. Moreover, \adaptGM{} is 43.4	and 18.05 minutes faster than \initializedFEM{} and \vesselGM{}, respectively. 
\adaptGM{} is about 12 times faster than \initializedFEM{} and significantly improves the RE.
While the RE difference in between the compliance methods is small, \adaptGM{} is much faster than \vesselGM{}.} 


\begin{table}[H] 
	\centering 
	\caption{Matching results for the BGM, VCGM and ACGM with a rigid initalialization as deformation reference.} 
	\label{Table:RealDataTRE} 
	\begin{adjustbox}{max width=\textwidth}
		\begin{tabular}{|c|cccc|} 
			\toprule
			\CoB{Porcine} & Matching & Number   &   RE [mm]  &   time  \\ 
			\CoB{dataset} & Method   &of matches&   \metric   &   [min] \\ 
			\midrule
			\multirow{2}{*}{PA  }
			& iGPR$_{Rigid}$  & 6 & 9.46$\pm$11.93 (65.3) & 2.1 \\ 
			& \initializedFEM{}  & 20 & 5.38$\pm$9.76 (58.4) &  39.3 \\ 
			37 landmarks & \vesselGM{} & 22 & 4.63$\pm$4.76 (20.8) & 8.5  \\ 
			& \adaptGM{} & 22 & 4.67$\pm$4.76 (21.1)  &  1.1 \\ 
			\midrule
			\multirow{2}{*}{PB}
			& iGPR$_{Rigid}$  & 4 & 13.71$\pm$8.10 (33.6) &  0.42 \\
			& \initializedFEM{}  & 11 & 7.61$\pm$7.34 (25.1) &  55.6 \\ 
			22 landmarks  & \vesselGM{} & 15 & 4.84$\pm$3.56 (15.3) &  35.7  \\ 
			& \adaptGM{} & 15 & 3.71$\pm$2.23 (7.6) & 7.0 \\ 
			\bottomrule
		\end{tabular}
	\end{adjustbox}
\end{table}

\subsubsection{Influence of the segmentation of the vessels}
\label{ss:res:RealData:SegInflu}
Different pre-processing parameters are used to obtain four sets of graphs from a partial side of the \textit{PA} dataset. Since only a partial side was used, the RE was evaluated only in 28 landmarks (two inserted tumors, 13 distinctive points and 13 bifurcations).
Using these graphs sets, the influence of the pre-processing steps on the matching methods is studied. 

As before, the optimal Mahalanobis threshold per method was searched from ten different increasing thresholds matchings. 
The second part of Table~\ref{Table:matchingComparison3GraphsRE} shows the optimal result of each method. 
For a majority of experiments, \adaptGM{} was faster, maintaining or even improving the RE. 
Only in experiment C (4th column of Table~\ref{Table:matchingComparison3GraphsRE}), the \adaptGM{} matching time was about the same as for \initializedFEM{}.
However, here the \adaptGM{} maximum RE is 3.4 mm better than for \initializedFEM{}.




The third part of Table \ref{Table:matchingComparison3GraphsRE} shows the results obtained with the highest  Mahalanobis thresholds used.
From the optimal threshold to the highest used, \initializedFEM{} increases the 4 segmentations matching time an average of 28.3 minutes. 
While, the \adaptGM{} average matching time only increases 6.6 minutes.

The fourth part of Table \ref{Table:matchingComparison3GraphsRE} shows the results using the $U_{TH}$ parameters to set the Mahalanobis thresholds. 
\vesselGM{} needs higher matching time than \adaptGM{}, up to the point that \vesselGM{} did not complete experiments C and D (marked with '-') as they exhausted the available RAM memory. 
\adaptGM{} using the $U_{TH}$ threshold selection is faster than the optimal $MC_{TH}$, the average for the 4 segmentations is 14.6 minutes faster.
While, the $U_{TH}$ selection average RE is only 0.2\,mm larger than the optimal MC threshold. 


\begin{table}[H] 
	\centering 
	\caption{Matching results evaluated on 28 landmarks of four different preprocessed graphs (obtained from real porcine data PA). }
	\label{Table:matchingComparison3GraphsRE} 
	\begin{adjustbox}{max width=\textwidth}
		\begin{tabular}{|c|ccc|ccc|ccc|ccc|}
			\toprule 
			&\multicolumn{3}{c|}{ A \ \ \ \sBifN{}=28, \tBifN{}=27}&\multicolumn{3}{c|}{ B  \ \ \ \sBifN{}=40, \tBifN{}=27}&\multicolumn{3}{c|}{ C  \ \ \ \sBifN{}=40, \tBifN{}=33}& \multicolumn{3}{c|}{ D  \ \ \ \sBifN{}=40, \tBifN{}=34} \\
			match   & Mah &   RE [mm]   &    time    & Mah &   RE [mm]   &    time     & Mah &   RE [mm]   &    time    & Mah &   RE [mm]   &    time       \\
			Func    & Th  &  $\mu\pm\sigma$ (max)  &  {[}min{]} & Th &  $\mu\pm\sigma$ (max)  &  {[}min{]} & Th &  $\mu\pm\sigma$ (max)  &  {[}min{]} & Th &  $\mu\pm\sigma$ (max)  &  {[}min{]}      \\[1.5ex]
			\multicolumn{13}{|l|}{optimal Mahalanobis threshold}
			\\\midrule
			BGM   & 1.6     & 2.9$\pm$ 1.7 (8.2) &  7.6 & 1.6 &  3.3$\pm$ 2.2 (7.9) & 10.1& 1.4 &  5.5$\pm$ 4.1 (15.2) &  31.3 & 1.8 &  3.1$\pm$ 2.3 (8.4) &  98.2  \\ 
			VCGM  & 4.4     &  2.9$\pm$ 1.7 (8.3) & 6.3 & 4.5 &  3.3$\pm$ 2.3 (8.4) &  5.0& 3.4 &  5.4$\pm$ 4.0 (14.9) &  6.2  & 4.2 &  3.0$\pm$ 2.4 (8.1) &  32.4  \\ 
			ACGM  & 3.9-6.6 & 2.9$\pm$ 1.7 (8.3) &  3.6 & 4.2-6.2 &  3.3$\pm$ 2.2 (7.3) &  4.7 & 3.9-4.8 &  5.1$\pm$ 3.4 (12.6) & 32.7  & 4.2-6.2 &  2.9$\pm$ 2.2 (7.4) &  42.4  
			\\[1.5ex]
			\multicolumn{13}{|l|}{Highest Mahalanobis threshold}
			\\\midrule
			BGM   & 2.8     & 2.9$\pm$ 1.7 (8.2) &  27.9 & 2.4 &  3.3$\pm$ 2.2 (7.9) & 50.4& 1.7 &  5.5$\pm$ 3.9 (15.2) &  84.0 & 1.8 &  3.1$\pm$ 2.3 (8.4) &  98.2  \\ 
			VCGM  & 7.0     &  2.9$\pm$ 1.7 (8.3) & 21.9 & 6.4 &  3.3$\pm$ 2.3 (8.4) &  18.9& 4.1 &  5.5$\pm$ 3.9 (14.8) &  27.6  & 4.5 &  3.0$\pm$ 2.4 (8.1) &  45.6  \\ 
			ACGM  & 5.0-7.2 & 2.9$\pm$ 1.7 (8.3) &  8.6 & 5.0-7.2 &  3.3$\pm$ 2.2 (7.3) &  12.6 & 3.9-4.8 &  5.1$\pm$ 3.4 (12.6) & 32.7  & 4.5-6.6 &  2.9$\pm$ 2.2 (7.4) &  55.8  
			\\[1.5ex]
			\multicolumn{13}{|l|}{Set MC threshold $U_{TH}$=16 mm, $U_{High}$=17 mm, and $U_{Low}$ 9mm}
			\\\midrule
			VCGM  & 4.8     &  2.9$\pm$ 1.7 (8.3) & 6.3 & 4.9 &  3.3$\pm$ 2.4 (8.5) &  10.0 & 5.8 & -$\pm$ - (-) &  -  & 4.9 &  -$\pm$ - (-) &  -  \\ 
			ACGM  & 2.7-5.1 & 2.9$\pm$ 1.7 (8.3) &  0.6 & 2.8-5.2 &  3.4$\pm$ 2.4 (8.6) &  1.24 & 3.3-6.2 &  5.5$\pm$3.9 (14.8) & 17.3  & 2.8-5.2 &  3.3$\pm$ 2.1 (8.5) &  5.8  
			\\\bottomrule
\end{tabular}
	\end{adjustbox}
\end{table}

\subsubsection{Influence of geodesic distance accuracy}
\label{ss:res:RealData:GeoInflu}

\begin{figure}[H]
	\captionsetup[subfigure]{aboveskip=-1pt,belowskip=4pt}
	\centering
		\adjincludegraphics[width=1.0\linewidth, trim={0 0 0 0},clip]{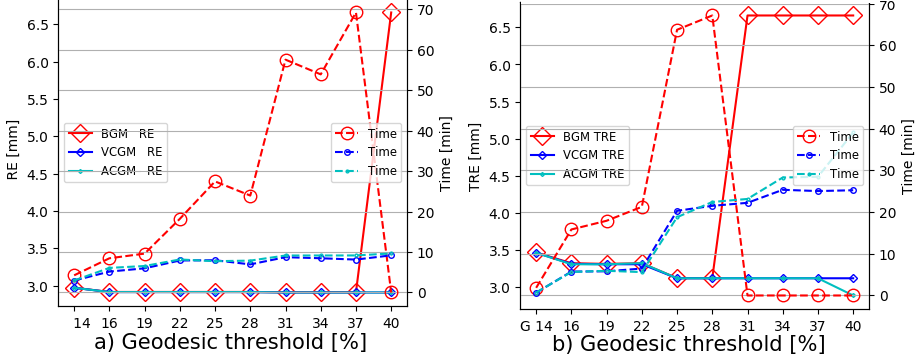} 
	\caption{
		Geodesic dependence of the RE and matching time on two different preprocessed graphs. 
		\initializedFEM{} increases drastically the matching time with higher geodesic threshold ($G_{TH})$. 
		While, the compliance methods depend less on the geodesic constraint.}
	\label{fig:GeodesicIncreaseS3Partial3}
\end{figure}

Using the graphs from previous experiments A and B, ten matchings are done to evaluate the geodesic influence.
Fig.\,\ref{fig:GeodesicIncreaseS3Partial3} shows the results using increasing geodesic $(G_{TH})$ and constant Mahalanobis thresholds. 
When $G_{TH}$ is increased, the \vesselGM{} and \adaptGM{} methods do not increase the search time as much as \initializedFEM{}.
 
Due to image noise and pre-processing, the geodesic distance is inaccurate in real data.
These inaccuracies represent a problem for \initializedFEM{}, because a small $G_{TH}$ does not allow to match inaccurate geodesic bifurcations (e.g. when $G_{TH}$<25\% in Fig. \ref{fig:GeodesicIncreaseS3Partial3}.b). 
An important advantage of the compliance methods is that the dependence on the geodesic constraint is reduced. 
This allows to use $G_{TH}=40\%$ with \adaptGM{}, which in experiment B finds an extra correct match and reduces the RE to 2.9 mm.


\section{Discussion}

\todo[inline]{In Figure 5, despite ACGM is explained as a better method than VCGM, ACGM appears to have a higher sensitivity to medium noise level than VCGM. Can the authors explain why?}
\todo[inline]{Moreover, the same paragraph suggests that unless UTHmin excludes, false matches can be introduced. This sounds arbitrary and a better justification is required.}

\JG{
In the 50\% noise case of Section \ref{ss:res:synth:optMah}, \adaptGM{} is more sensitive to noise than \vesselGM{} because $MC_{min}$ was set without using Eq. \ref{eq:MahalanobisComplianceThreshold}. Setting a correct $MC_{min}$ is hard in the presence of noise that cannot be filter with the geodesic constraint. 
This specific situation is an \adaptGM{}'s limitation that was overcome when Eq. \ref{eq:MahalanobisComplianceThreshold} defined $MC_{min}$ in all other experiments.
}

The experiment in Section \ref{s:res:RealData} shows with real data that \adaptGM{} is faster and more accurate than \initializedFEM{}. Globally, \adaptGM{} correctly registers the two trees, except for few leaf vessels in the upper middle part of Fig.\,\ref{fig:matchFineLarge}c.
These few vessels are not well transformed because the corresponding bifurcations are missing in the target segmentation.
These errors can be solved after the coarse bifurcation matching presented here, either by improving the fine matching of high deformation edges or using vessels' end points.

\todo[inline]{compared ACGM with BGM. Why authors do not refer VCGM method as a comparison with ACGM and/or BGM? Did you assume, at this stage, that ACGM is better than VCGM (also implemented)? If you assumed that, it must be explained very well, previously, in other sections.}

\todo[inline]{Uth all together dependency to scale all together? continue in sec 2.2.1} 
\CoB{
	
It is hard to directly set the optimal Mahalanobis compliance thresholds.	
In \vesselGM{}, $U_{TH}$ removes the initialization and FE model dependency. However, it still depends on the deformation magnitude, incremental hypotheses generation and scale. A large $U_{TH}$ finds a correct match but increases the matching time. 
\adaptGM{} allows to overestimate $U_{High}$ to guarantee the correct solution and thanks to the rigid-to-soft approach it increases minimally the matching time. This allowed to set an approximate range that matched efficiently a wide variety of experiments. That's why \adaptGM{} is considered better than \vesselGM{}.	
}	

\JG{The proposed \adaptGM{} method is up to one order of magnitude faster than \initializedFEM{}. 
Besides being faster, \adaptGM{} maintains or even improves the RE in real and synthetic experiments.
All the parameters of the algorithm, including the incremental maximum displacement range, are easy to set and can work correctly in a wide range of scenarios. }
The method allows to augment anatomical structures non-visible in the intra-operative image. 
This improves the visualization of the target and risk anatomy, which is a very important clinical need that could, for example, allow surgeons to reach small structures during image guided procedures.   

\adaptGM{} was successful even under very large deformations while using an automatic segmentation method on the porcine liver dataset. This robustness was also demonstrated by adding several levels of noise and deformations in the synthetic datasets.
The efficient \adaptGM{} method handles a large number of noise bifurcations. 
Still maintaining the matching time within acceptable intra-operative constraints. 
This alleviates the need of low noise intra-operative automatic image segmentation. 
The compliance filtering reduces the dependence on geodesic constraints, this could even allow to efficiently match unconnected graphs. 


\CO{ 
These results are promising but further evaluation is required to validate the in-vivo applicability. 
First, the impact of different image acquisition conditions (resolution, contrast injection or artifacts) should be studied. 
Second, experiments on larger human datasets are required to prove the clinical importance. 
Although the clinical set-up is relatively common, collecting a large dataset with expert manual ground truth (to quantify registration error) requires significant amount of work.}

The experiments show that \adaptGM{} is close to accurately match unconnected, very noisy and highly deformed graphs. Thus, a faster implementation of \adaptGM{} (with more efficient mechanical models, artificial intelligence matching methods, and/or parallel implementations) can allow future US-CT registration studies.  

\section{Acknowledgments} 

The authors are grateful for the support from Inria, the MIMESIS and MAGRIT teams, and IHU Strasbourg. 
Jaime Garcia Guevara is supported by the Grand Est region and Inria.

\section{CONFLICT OF INTEREST}
\CO{We declare that this article is free from conflicts of interest.}

\bibliographystyle{abme}
\bibliography{abme2018.bib}

\end{document}